\documentclass[sigconf]{acmart}
\usepackage{colortbl}

\usepackage{multirow}
\usepackage{tabularx}
\usepackage{arydshln}
\usepackage{adjustbox}
\usepackage{array}
\AtBeginDocument{%
  \providecommand\BibTeX{{%
    \normalfont B\kern-0.5em{\scshape i\kern-0.25em b}\kern-0.8em\TeX}}}

\setcopyright{acmcopyright}
\copyrightyear{2018}
\acmYear{2018}
\acmDOI{XXXXXXX.XXXXXXX}

\acmConference[Conference acronym 'XX]{Make sure to enter the correct
  conference title from your rights confirmation emai}{June 03--05,
  2018}{Woodstock, NY}
\acmPrice{15.00}
\acmISBN{978-1-4503-XXXX-X/18/06}

\begin{document}

\title{CoAScore: Chain-of-Aspects Prompting for NLG Evaluation}

\author{Peiyuan Gong}
\affiliation{%
  \institution{GSAI, Renmin University of China}
  \city{Beijing}
  \country{China}}
\email{pygongnlp@gmail.com}

\author{Jiaxin Mao}
\affiliation{%
  \institution{GSAI, Renmin University of China}
  \city{Beijing}
  \country{China}}
\email{maojiaxin@gmail.com
}


\begin{abstract}
Recently, natural language generation (NLG) evaluation has shifted from a single-aspect to a multi-aspect paradigm, allowing for a more accurate assessment.
Large language models (LLMs) achieve superior performance on various NLG evaluation tasks. 
However, current work often employs the LLM to independently evaluate different aspects, which largely ignores the rich correlation between various aspects. 
To fill this research gap, in this work, we propose an NLG evaluation metric called CoAScore. 
Powered by LLMs, the CoAScore utilizes multi-aspect knowledge through a CoA (\textbf{C}hain-\textbf{o}f-\textbf{A}spects) prompting framework when assessing the quality of a certain aspect.
Specifically, for a given aspect to evaluate, we first prompt the LLM to generate a chain of aspects that are relevant to the target aspect and could be useful for the evaluation.
We then collect evaluation scores for each generated aspect, and finally, leverage the knowledge of these aspects to improve the evaluation of the target aspect.
We evaluate CoAScore across five NLG evaluation tasks (e.g., summarization, dialog response generation, etc) and nine aspects (e.g., overall quality, relevance, coherence, etc). 
Our experimental findings highlight that, in comparison to individual aspect evaluation, CoAScore exhibits a higher correlation with human judgments. This improvement significantly outperforms existing unsupervised evaluation metrics, whether for assessing overall quality or other aspects.
We also conducted extensive ablation studies to validate the effectiveness of the three stages within the CoAScore framework and conducted case studies to show how the LLM performs in these stages. Our code and scripts are available.
\end{abstract}






\received{20 February 2007}
\received[revised]{12 March 2009}
\received[accepted]{5 June 2009}

\maketitle

\section{Introduction}
Natural language generation (NLG) evaluation holds a pivotal and essential role within the domain of natural language processing research \cite{sai2022survey}. It focuses on measuring the quality of system-generated hypotheses (e.g. summaries, responses) based on provided sources (e.g. articles, conversations) for various NLG tasks, such as text summarization \cite{fabbri2021summeval, durmus2020feqa}, dialog response generation \cite{mehri2020usr, mehri2020unsupervised, sinha2020learning}, machine translation \cite{rei2020comet, zhang2019bertscore, sellam2020bleurt}, and so on.
Besides, these evaluation metrics also play a vital role in enabling researchers and developers to analyze the strengths and weaknesses of their NLG models. 
Nonetheless, it's essential to acknowledge that evaluating NLG systems is an intricate and demanding task. As the NLG field continues to advance and systems become more complex and capable, the quest for evaluations that are not only reliable but also interpretable and comprehensive gains even more significance.

In response to these challenges, there has been a concerted effort to develop automatic NLG evaluation metrics that do not require manual annotations.
BLEU \cite{papineni2002bleu}, ROUGE \cite{lin2004rouge}, and METEOR \cite{banerjee2005meteor} represent a category of rule-based evaluation metrics. These metrics rely on a set of heuristic criteria and rules to aid in the assessment of system-generated hypotheses. 
Benefiting from the advancements in pre-training techniques \cite{devlin2018bert, lewis2019bart}, BERTScore \cite{zhang2019bertscore}, BARTScore \cite{yuan2021bartscore}, and UniEval \cite{zhong2022towards} harness pre-training knowledge to augment the evaluation process.
Furthermore, recognizing the formidable comprehension and reasoning capabilities of large language models (LLMs) \cite{zhao2023survey}, the NLP community has increasingly turned its attention to LLM-based evaluation of NLG systems \cite{liu2023gpteval, fu2023gptscore, liu2023evaluate}. This approach has garnered significant interest and traction.

In recent years, there has been a shift from solely measuring a single aspect to emphasizing the evaluation of multiple aspects in NLG systems \cite{sai2022survey}. These aspects may include engagement, fluency, relevance, and more. Notably, it has been observed that the scores assigned to different aspects can significantly vary. This underscores the suboptimal and risky nature of exclusively assessing the overall quality of generated text in NLG.
With the growing availability of multi-aspect evaluation datasets like SummEval \cite{fabbri2021summeval} and TopicalChat \cite{mehri2020usr}, training-based evaluation metrics have gained importance for evaluating various aspects of NLG systems. For instance, USR \cite{mehri2020usr} employs models trained on distinct corpora tailored to different aspects, while UniEval \cite{zhong2022towards} facilitates multi-aspect evaluation through mixed corpus training.
Moreover, LLM-based evaluation metrics, such as GPTScore \cite{fu2023gptscore} and GPTEval \cite{liu2023gpteval}, leverage their robust ability to follow instructions to achieve more precise multi-aspect text evaluation without requiring supervised data. These metrics explicitly declare and describe the aspects to be evaluated within the instructions, enhancing the evaluation process.

Given that sentences generated by NLG systems often cover various aspects, a single aspect can frequently have associations with several additional aspects.
\citet{novikova2018rankme} illustrates that human evaluations often display a certain level of correlation among scores across different evaluation aspects, encompassing informativeness, naturalness, and quality.
\citet{madaan2023self} demonstrates that when refining overall output quality through LLM feedback, deconstructing the overall feedback into multiple components can further amplify the effectiveness of this mechanism in refining sentence generation.
\citet{tevet2020evaluating} establishes that when appraising the diversity of outputs from NLG systems, it can be further segmented into distinct diversity categories, covering both content and form. Moreover, it can be further partitioned to form a tree-like structure.
However, even though NLG evaluation has shifted from a singular to a multi-aspect approach, a significant gap persists. Individual aspects such as fluency, coherence, and relevance are frequently evaluated independently, neglecting the complex interrelationships among them. An integrated evaluation framework that recognizes these interdependencies, such as the close link between fluency and coherence or relevance and informativeness, is essential for a deeper comprehension of the capabilities and constraints of NLG systems

To tackle this problem, we introduce CoAScore, an LLM-based evaluation metric that employs a chain-of-aspects prompting framework to assess the quality of system-generated hypotheses with respect to the target aspect. By leveraging relevant aspects as reference points, it enriches the evaluation process in a chain-of-thought manner, resulting in more precise assessments of the target aspect.
In detail, when evaluating a specific aspect, we first generate a chain of relevant aspects for reference, closely related to the aspect under evaluation. These aspects are pre-scored for quality. Finally, we integrate this knowledge, including descriptions and scores of relevant aspects, to enhance LLM's capacity in evaluating NLG systems.
To validate the effectiveness of our approach, we conducted experiments on nine aspects (e.g., overall quality, relevance, coherence, etc) across five diverse evaluation datasets (e.g., summarization, dialog response generation, etc). The results demonstrate that CoAScore exhibits a stronger correlation with human judgments compared to isolated aspect evaluations, surpassing existing unsupervised baselines in assessing both overall quality and other aspects. Moreover, as the number of relevant aspects increases, the correlation between CoAScore and human judgments often becomes stronger.
As CoAScore comprises multiple stages based on LLM, we conducted several comparative experiments to individually verify the effectiveness of each stage within the CoAScore framework. Additionally, we constructed case studies to provide further explanations and insights.
% To address the above problem, we propose an LLM-based evaluation metric, named CoAScore, which employs a chain-of-aspects prompting framework to assess the given aspect's quality of system-generated hypotheses.
% Incorporating the knowledge of relevant aspects as points of reference, it amplifies the capacity of the evaluation model in a chain-of-thought manner, resulting in a more accurate assessment of the target aspect.
% In detail, when evaluating a certain aspect, we first generate multiple relevant aspects for reference. These generated aspects are closely related to the target aspect being evaluated. 
% Then, we score the quality of these generated aspects beforehand. 
% Finally, by taking into account the knowledge (descriptions and scores) of relevant aspects to enhance the ability of LLM to evaluate NLG systems.
% To validate the efficacy of our approach, we perform experiments across 9 aspects within 5 distinct evaluation datasets.
% Empirical findings reveal that CoAScore exhibits a stronger correlation with human judgments when compared to isolated aspect evaluations. This correlation outperforms existing unsupervised baselines in appraising both overall quality and other aspects.
% We verified the importance of multi-aspect knowledge in CoAScore through ablation experiments and show the reasons behind CoAScore's superiority through case studies.

Our \textbf{main contributions} are listed as follows:
\begin{itemize}
    \item We introduce a novel evaluation metric called CoAScore, which utilizes a chain-of-aspects prompting framework. This framework employs LLM to create, measure, combine, and leverage a sequence of relevant aspects as points of reference, aiming to enhance the evaluation capability for the specific target aspect.
    \item Experiment results demonstrate that CoAScore correlates best with human judgments across various NLG evaluation tasks, not only in terms of overall quality but also in other aspects, respectively. 
    \item We carry out comprehensive ablation investigations to confirm the efficiency of the three stages incorporated in the CoAScore framework and also conduct case studies to illustrate how the LLM works during these stages.
\end{itemize}

\section{Related Work}
% \subsection{NLG Metrics}
The rapid expansion of NLG systems has underscored the vital need for robust and user-friendly evaluation metrics to assess the quality of text generated by these systems. The varied and complex nature of NLG applications, spanning from translation to chatbots, demands precise evaluation criteria. Consequently, researchers and practitioners have dedicated substantial efforts to enhance and standardize NLG evaluation techniques, making it a pivotal area of contemporary research. This focus is aimed at fostering the creation of more efficient and trustworthy NLG systems. We categorize NLG evaluation methods into three distinct types:

\textbf{Rule-based metrics} are a class of evaluation methods that assess the quality of hypotheses by employing predefined heuristic rules.
BLEU \cite{papineni2002bleu} utilizes heuristic rules such as n-gram matching to measure the similarity between references and hypotheses.
ROUGE \cite{lin2004rouge} employs techniques including n-gram matching, lemmatization, and part-of-speech tagging to evaluate the similarity between each sentence pair.
METEOR \cite{banerjee2005meteor} leverages various unigram matching algorithms to quantify the resemblance between references and hypotheses.
These rule-based metrics serve as essential tools in the NLG evaluation toolkit, providing a foundation for assessing the quality of generated text using well-defined rules and criteria.

\textbf{Machine-learned metrics} represent a category of NLG evaluation methods that leverage the potential of pre-trained knowledge to enhance their evaluation judgments. 
BERTScore \cite{zhang2019bertscore} computes similarity scores between tokens in the hypothesis and reference using pre-trained token embeddings. 
BARTScore \cite{yuan2021bartscore} suggests that a high-quality hypothesis should be effortlessly generated from the source or reference using pre-trained models.
COMET \cite{rei2020comet} proposes two cross-language machine translation metrics, which employ pre-train model to form estimator and translation ranking frameworks respectively.
DEB \cite{sai2020improving} is a BERT-based dialog metric pre-trained on a large size of Reddit conversations and fine-tune on a multi-references dialog dataset.
PT-M$^2$ \cite{gong2022revisiting} as a grammatical error correction metric, leveraging pre-trained knowledge to measure the importance of different edits for correcting sentences with grammatical errors.

\begin{figure*}[t]
\centering
\includegraphics[width=\textwidth]{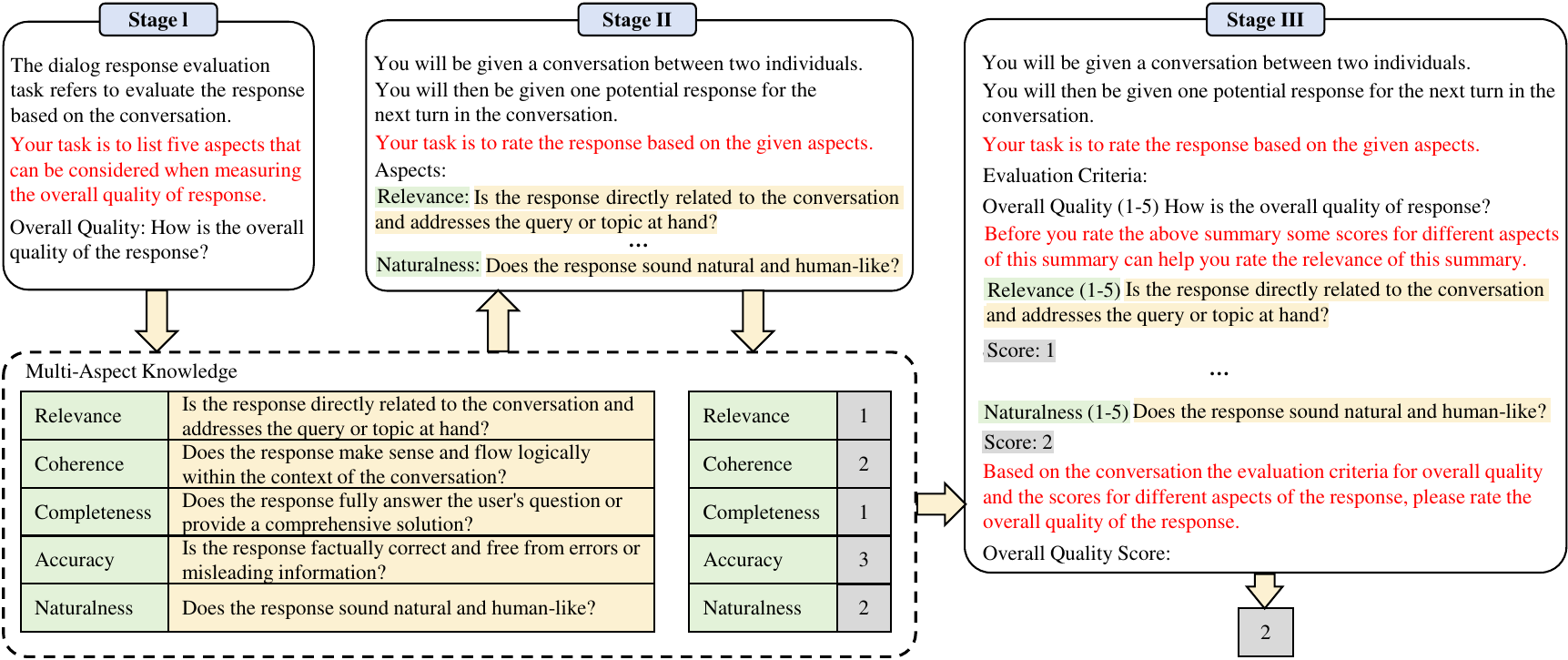}
\caption{
    The overall prompting framework of CoAScore.  Given the evaluation task instruction $\boldsymbol{t}$, evaluation aspect $a$, source $\boldsymbol{s}$ and hypothesis $\boldsymbol{h}$, CoAScore needs to measure the quality of the hypothesis in that aspect. CoAScore consists of three distinct stages, and each stage is carried out by LLM: \textbf{(I)} Generating a chain of aspects that will be used as references when evaluating the target aspect. These generated aspects are chosen to be closely related to the target aspect; \textbf{(II)} Scoring each of the generated aspects for the hypothesis; \textbf{(III)} Leveraging the knowledge about the chain of relevant aspects to enhance the evaluation capability for the specific target aspect. Some detailed information, such as conversation context and replies, has been omitted in the prompts and can be found in Appendix A.
}
\label{coascore}
\end{figure*}

\textbf{LLM-based metrics} harness the impressive text comprehension and reasoning capabilities of Large Language Models (LLMs) for evaluation purposes, contributing to more advanced and context-aware NLG assessments.
GPTScore \cite{fu2023gptscore} attempts to achieve multi-faceted evaluation through instructions without training corpus.
G-Eval \cite{liu2023gpteval} leverages LLMs with chain-of-thoughts and a form-filling paradigm to evaluate the quality of NLG systems.
InstructScore \cite{xu2023instructscore} is an explainable evaluation metric, employing LLMs to generate diagnostic reports for system-generated hypotheses.
Moreover, our findings reveal that varying prompts often lead to relatively significant differences in the evaluation results \cite{chen2023exploring, wang2023chatgpt, kocmi2023large, huynh2023understanding}.
Additionally, there is ongoing research into leveraging LLMs for the generation of NLG evaluation datasets, aiming to reduce the need for manual evaluations to some extent \cite{mohtashami2023learning, chiang2023can}.
In our work, we utilize LLMs to generate a chain of aspects as references for aspect assessment. These aspects are scored beforehand, and we leverage the knowledge derived from them to enhance the ability to evaluate the target aspect. This approach showcases the versatility and potential of LLMs in improving NLG evaluation methodologies.

\section{Methodology}
In this section, we provide an overview and delve into the details of CoAScore. 
In the realm of NLG evaluation task, each aspect can exhibit connections to multiple other aspects \cite{zhong2022towards} or be decomposed into several relevant sub-aspects \cite{tevet2020evaluating}.
However, the evaluation metrics devised in most research predominantly focus on modeling each specific evaluation aspect independently, overlooking the interconnections between different aspects \cite{fu2023gptscore, liu2023gpteval, xu2023instructscore}. To address the above problem, CoAScore adopts a chain-of-aspects prompting framework, where it takes into account the knowledge of relevant aspects when evaluating a given aspect to make better decision.
As shown in Figure \ref{coascore}, at the beginning stage, CoAScore generates a chain of relevant aspects that can aid in evaluating a specific target aspect of the hypothesis. 
Subsequently, at stage II, each generated aspect is scored.
Finally, at the last stage, the knowledge of relevant aspects (definitions and scores) is leveraged to facilitate the evaluation of the target aspect. 
This section is structured into four parts. The initial part presents the problem definition of the NLG evaluation task. The subsequent three parts correspond to the implementation of the three stages of CoAScore.

\subsection{Problem Definition}
In the context of NLG evaluation, the evaluation of the hypothesis $\boldsymbol{h}$ is based on a particular aspect $a$ (e.g., overall quality, relevance, fluency, etc.), considering a source sentence $\boldsymbol{s}$ and a reference sentence $\boldsymbol{r}$ \cite{sai2022survey, celikyilmaz2020evaluation}. The main objective is to develop an aspect-aware evaluation function $f_a$ capable of accurately scoring the hypothesis with respect to the specified aspect:

\begin{equation}
    y_{a} = f_a(\boldsymbol{h}, \boldsymbol{s}, \boldsymbol{r})
\end{equation}

Where $y_a$ means the score of the hypothesis on the target aspect. 
% An important observation is that reference-based metrics encounter one-to-many problem in various NLG evaluation tasks\cite{zhao2017learning}. Given the impressive context understanding abilities of LLMs, they seem better suited for reference-less evaluation paradigms \cite{liu2023gpteval}. Consequently, 
To avoid the one-to-many problem in reference-based metrics \cite{chan2021enhancing, ji2022achieving, shi2023rade}, in our work, CoAScore is introduced as a reference-less evaluation metric, represented as $y_{a} = f_a(\boldsymbol{h}, \boldsymbol{s})$.

% \subsection{Baseline}
% In this part we describe the baseline for leveraging LLMs to evaluate. Given the task description $\boldsymbol{t}$, the aspect $a$ used for evaluation, source sentence $\boldsymbol{s}$ and hypothesis sentence $\boldsymbol{h}$, we need to calculate the score $y_a$ of hypothesis$h$ in terms of aspect $a$. Here is an example about how to work with a certain prompt:

% \begin{tcolorbox}[colback=blue!5!white,colframe=blue!75!black]
%  Based on the news article and the evaluation criteria for relevance, please rate the relevance of the summary
% \end{tcolorbox}

\subsection{Relevant Aspect Generation}
\label{rag}
To ensure that CoAScore can employ knowledge of relevant aspects to assist evaluation, the initial step in our approach involves generating a chain of relevant aspects associated with the under-evaluated aspect, as shown in Figure \ref{coascore}. These aspects will serve as references during the evaluation of the target aspect.
Specifically, given the task description $\boldsymbol{t_{rag}}$ and the aspect $a$ used for evaluation, the first step we have to do is to generate $m$ relevant aspects that can be taken into count when evaluating the target aspect. The generated aspects can be formulated as:

\begin{equation}
    \boldsymbol{A}=A_1, A_2, \ldots, A_m, \quad A_i = \{n_i: d_i\}
\end{equation}

Where $\boldsymbol{A}$ refers to the chain of aspects related to the given aspect $a$. Each $A_i$ is represented in dictionary format, with the aspect name $n_i$ (e.g. relevance, coherence) as the key and a detailed description $d_i$ of the aspect as the value.

\subsection{Relevant Aspect Scoring}
Once a chain of relevant aspects for reference has been generated, the subsequent step entails scoring the quality of each one in relation to the hypothesis.
These scores serve as significant references to assist in the final evaluation.
It is essential to emphasize that, for efficiently obtaining the evaluation result of the target aspect $a$, we generate scores for multiple relevant aspects simultaneously at this stage, as depicted in Figure \ref{coascore}.
In detail, given the task description $\boldsymbol{t_{ras}}$, source sentence $\boldsymbol{s}$, hypothesis sentence $\boldsymbol{h}$, and the chain of aspects $\boldsymbol{A}$ observed from the previous stage, our objective is to score each $A_i$ from $\boldsymbol{A}$ for the hypothesis $\boldsymbol{h}$. Scores for the chain of relevant aspects can be formulated as:

\begin{equation}
    \boldsymbol{S}=S_1, S_2, \ldots, S_m, \quad S_i = \{n_i: s_i\}
\end{equation}

Here, $\boldsymbol{S}$ represents the scores of the hypothesis on the relevant aspects $\boldsymbol{A}$, where $S_i$ is in one-to-one correspondence with $A_i$. Each $S_i$ is presented in dictionary format, by using the aspect name $n_i$ as the key and associating it with the respective score $s_i$.

\subsection{Chain-of-Aspects Scoring}
Upon acquiring the definitions of relevant aspects and their corresponding scores pertaining to the hypothesis, the next step encompasses reassembling them to fully leverage their role at this stage:

\begin{equation}
    \boldsymbol{K}=K_{1}, K_{2}, \ldots, K_{m}, \quad K_{i} = \{n_i: (d_i, s_i)\}
\end{equation}

Where $\boldsymbol{K}$ represents knowledge about the chain of relevant aspects, aiding in the evaluation of the target aspect $a$. Each $K_{i}$ is represented as a dictionary, where the name of the relevant aspect $n_i$ serves as the key, and the description $d_i$ along with the corresponding score $s_i$ form a tuple $(d_i, s_i)$ as the value.

Once we have acquired comprehensive knowledge about the chain of aspects that are relevant to the evaluation of a specific aspect, we utilize this information to assist in the final evaluation process. 
In summary, our current work focuses on the complete process of measuring the target aspect using the chain-of-aspects knowledge. Given the Chain-of-Aspects Scoring task description, denoted as $\boldsymbol{t_{coa}}$, we have the aspect $a$ for evaluation, along with the source sentence $\boldsymbol{s}$ and the hypothesis sentence $\boldsymbol{h}$. Additionally, we possess the chain-of-aspects knowledge represented as $\boldsymbol{K}$, where each $K_i$ consists of the aspect description $d_i$ and its corresponding score $s_i$. The final step involves utilizing the aforementioned information to compute the score $y_a$ associated with the hypothesis $\boldsymbol{h}$ for the given aspect $a$, as depicted in Figure \ref{coascore}.
Please see Appendix A for the detailed prompts used in these three stages.

\section{Experiments} 
\subsection{Datasets}
\begin{table}[t]
\centering
\caption{Statistics of the used NLG evaluation datasets.}
\begin{adjustbox}{max width=\linewidth}
\begin{tabular}{lcc}
    \toprule
    \bf Dataset & \bf Size & \bf Aspects\\
    \midrule
    \multirow{2}{*}{SummEval} & \multirow{2}{*}{1,600} & \multicolumn{1}{c}{Coherence, Consistency}\\
    & & Fluency, Relevance \\
    \midrule
    \multirow{2}{*}{TopicalChat} & \multirow{2}{*}{300} & \multicolumn{1}{c}{Overall, Natural, Context}\\
    & & Understandable, Engaging \\
    \midrule
    \multicolumn{1}{l}{OpenMEVA} & 1,000 & \multirow{3}{*}{Overall}   \\
    \multicolumn{1}{l}{BAGEL} & 404 &   \\
    \multicolumn{1}{l}{IWSLT14} & 1,000 & \\
    \bottomrule
\end{tabular}
\end{adjustbox}
\label{datasets}
\end{table}

\begin{table*}[t]
\centering
\caption{Correlations between metrics and human judgments regarding the overall quality across four NLG evaluation tasks. Our proposed CoAScore exhibits stronger correlations with human judgments across all these tasks. We highlight the \textbf{highest} score in bold and the \underline{second-highest} score with underlines.}
\begin{adjustbox}{max width=\linewidth}
\begin{tabular}{lcccccccccccc}
\toprule
\multirow{2}{*}{\textbf{Metrics}} & 
\multicolumn{3}{c}{\textbf{OpenMEVA}} & \multicolumn{3}{c}{\textbf{BAGEL}} &
\multicolumn{3}{c}{\textbf{IWSLT14}} & \multicolumn{3}{c}{\textbf{TopicalChat}} \\
\cmidrule(lr){2-4} \cmidrule(lr){5-7} \cmidrule(lr){8-10} \cmidrule(lr){11-13}  
& $\gamma$ & $\rho$ & $\tau$ & $\gamma$ & $\rho$ & $\tau$ & $\gamma$ & $\rho$ & $\tau$ & $\gamma$ & $\rho$ & $\tau$ \\
\midrule 
\multicolumn{10}{l}{~~\textit{Rule-based metrics}} \\
BLEU & -0.006 & -0.012 & -0.009 & 0.063 & 0.022 & 0.016 & 0.328 & 0.316 & 0.220 & 0.228 & 0.293 & 0.205\\
ROUGE & -0.007 & -0.018 & -0.013 & 0.062 & 0.029 & 0.022 & 0.387 & 0.371 & 0.257 & 0.268 & 0.285 & 0.200\\
METEOR & 0.016 & 0.006 & 0.004 & 0.048 & 0.030 & 0.022 & 0.355 & 0.331 & 0.228 & 0.364 & 0.367 & 0.257\\
\midrule
\multicolumn{10}{l}{~~\textit{Machine-learned metrics}} \\
BERTScore & 0.279 & 0.251 & 0.176 & 0.147 & 0.096 & 0.069 & 0.561 & 0.567 & 0.408 & 0.294 & 0.321 & 0.222\\
BARTScore & 0.036 & 0.034 & 0.024 & 0.045 & 0.010 & 0.007 & 0.312 & 0.281 & 0.193 &  0.370 & 0.390 & 0.270\\
\midrule 
\multicolumn{10}{l}{~~\textit{Ours}} \\
LLMScore & 0.369 & 0.360 & 0.299 & 0.283 & 0.222 & 0.194 & 0.564 & 0.577 & 0.470 & 0.591 & 0.592 & 0.474\\
LLMScore$_{CoT}$ & 0.333 & 0.327 & 0.271 & 0.352 & \underline{0.262} & \underline{0.225} & 0.564 & 0.566 & 0.446 & 0.610 & 0.588 & 0.461\\
\hdashline
CoAScore$_{(n=5)}$ & 0.295 & 0.316 & 0.237 & \bf 0.409 & \bf 0.347 & \bf 0.272 & \bf 0.609 & \bf 0.621 & \bf 0.477 & 0.614 & 0.625 & \underline{0.476}\\
CoAScore$_{(n=10)}$ & \underline{0.408} & \bf 0.414 & \underline{0.304} & \underline{0.396} & 0.261 & 0.222 & \underline{0.595} & 0.607 & 0.466 & \underline{0.622} & \underline{0.638} & \bf 0.479\\
CoAScore$_{(n=20)}$ & \bf 0.414 & \underline{0.411} & \bf 0.315 & 0.393 & 0.254 & 0.205 & 0.592 & \underline{0.610} & \underline{0.474} & \bf 0.634 & \bf 0.641 & \bf 0.479\\ 
\bottomrule
\end{tabular}
\end{adjustbox}
\label{main_result}
\end{table*}

To validate our metric's effectiveness, we conduct experiments on five NLG evaluation datasets listed in Table \ref{datasets}. Initially, we assess our approach's capability in appraising the overall quality across these datasets:
\textbf{TopicalChat} \cite{mehri2020usr}: Evaluating the potential of evaluation metrics to effectively gauge response quality in dialog systems.
\textbf{OpenMEVA} \cite{mairesse2010phrase}: Measuring the evaluation effectiveness of metrics in story generation.
\textbf{BAGEL} \cite{guan2021openmeva}: Assessing the capacity of evaluation metrics to evaluate data-to-text systems that generate descriptions for provided tables.
\textbf{IWSLT14} \cite{kreutzer2018reliability}: Gauging the proficiency of evaluation metrics in evaluating the machine translation from German to English.

Beyond the overall quality assessment, we delve into CoAScore's versatility in evaluating various aspects. We proceed to assess the CoAScore framework across four aspects within the TopicalChat dataset: Natural (NAT), Understandable (UND), Interest (INT), and Maintains Context (CON).
Furthermore, our evaluation extends to incorporate the \textbf{SummEval} dataset \cite{fabbri2021summeval}, concentrating on four core aspects of summary evaluation: Coherence (COH), Consistency (CON), Fluency (FLU), and Relevance (REL).

\subsection{Baselines}
We compare CoAScore with the following common unsupervised NLG evaluation metrics: 
\textbf{BLEU} \cite{papineni2002bleu} employs heuristic rules like n-gram matching to assess the similarity between references and hypotheses. 
\textbf{ROUGE} \cite{lin2004rouge} utilizes techniques like n-gram matching, lemmatization, and part-of-speech tagging to gauge the similarity between sentence pairs.
\textbf{METEOR} \cite{banerjee2005meteor} employs diverse unigram matching algorithms to quantify the likeness between references and hypotheses.
\textbf{BERTScore} \cite{zhang2019bertscore} calculates token similarity between sentence pairs and employs a greedy matching approach to maximize the similarity scores.
\textbf{BARTScore} \cite{yuan2021bartscore} transform the evaluation task into a sequence generation problem, utilizing generation probabilities to gauge the quality of hypotheses. 
\textbf{LLMScore} serves as our self-developed LLM-based evaluation metric, used as a comparative baseline for CoAScore.
\textbf{LLMScore$_{CoT}$} is inspired by the Chain-of-Thought framework \cite{wei2022chain}, enhancing the evaluation ability of LLMScore by generating thinking processes through "Let's think step by step:".

\subsection{Implementation Details}
Our implementation involves utilizing BLEU \cite{papineni2002bleu}, ROUGE \cite{lin2004rouge}, and METEOR \cite{banerjee2005meteor} from the \textbf{evaluate} repository\footnote{https://github.com/huggingface/evaluate}. 
When utilizing BERTScore \cite{zhang2019bertscore}, we apply the \texttt{bert-base-uncased} pre-trained model \cite{devlin2018bert} and opt for F$_1$ as the ultimate score\footnote{https://github.com/Tiiiger/bert\_score}. 
Employing \texttt{bart-base} pre-train model \cite{lewis2019bart}, BARTScore \cite{yuan2021bartscore} generates the evaluation scores from hypotheses to references\footnote{https://github.com/neulab/BARTScore}. 
For LLMScore, LLMScore$_{CoT}$ and CoAScore, we harness ChatGPT\footnote{https://chat.openai.com/}, a widely recognized and extensively used large language model, with a configuration of “$\text{temperature}=0, \text{n}=1$”. 
In the context of CoAScore, we leverage 5, 10, and 20 relevant aspects to enhance the evaluation of each target aspect, referred to as CoAScore$_{(n=5)}$, CoAScore$_{(n=10)}$, and CoAScore$_{(n=20)}$ respectively.

\subsection{Metrics and Evaluation Strategy}
In this work, We employ three correlation coefficients to quantify the correlation scores between NLG evaluation metrics and human judgments:
\textbf{Pearson $\gamma$} \cite{mukaka2012guide} measures strength of linear relationship between two variables.
\textbf{Spearman $\rho$} \cite{zar2005spearman} evaluates the strength of monotonic association, accommodating nonlinear relationships.
\textbf{Kendall-Tau $\tau$} \cite{kendall1938new} measures the concordance and discordance in rankings, indicating the ordinal correlation between variables.

Regarding the evaluation strategy, we compute the correlation coefficients for the aforementioned metrics at the \textbf{Dataset-level}, which implies that correlation scores are derived from the NLG system outputs of all samples.

\subsection{Main Results}
We ascertain the efficacy of our evaluation metrics by contrasting them with rule-based metrics (BLEU \cite{papineni2002bleu}, ROUGE \cite{lin2004rouge}, and METEOR \cite{banerjee2005meteor}) and machine-learned metrics (BERTScore \cite{zhang2019bertscore} and BARTScore \cite{yuan2021bartscore}) across various NLG evaluation datasets. This comparison involves measuring the correlation between automatic evaluation metrics and human judgments on multiple aspects.

Table \ref{main_result} presents correlations between evaluation metrics and human judgments on overall quality across four evaluation datasets. Our proposed CoAScore demonstrates stronger correlations with human judgments compared to both rule-based metrics and machine-learned metrics. This trend is consistent in terms of Pearson, Spearman, and Kendall-Tau coefficients \cite{mukaka2012guide, zar2005spearman, kendall1938new}.
Furthermore, CoAScore consistently outperforms our self-designed LLM-based metrics, including LLMScore and LLMScore$_{CoT}$, in correlation with human judgments. This pattern persists across varying numbers of relevant aspects (5, 10, and 20 aspects).
Simultaneously, we observed that CoAScore's performance improves as the number of relevant aspects increases, particularly in the OpenMEVA \cite{guan2021openmeva} and TopicalChat datasets \cite{mehri2020usr}. 
Meanwhile, on the BAGEL \cite{mairesse2010phrase} and IWLST14 \cite{kreutzer2018reliability} datasets, employing just five relevant aspects as references, CoAScore exhibits notably larger correlation scores with human assessments compared to LLMScore's corresponding scores.
The aforementioned results demonstrate the effectiveness of CoAScore in leveraging multi-aspect knowledge for evaluation.

\begin{table}[t]
\centering
\caption{Spearman correlations between metrics and human judgments on four aspects of the SummEval. CoAScore demonstrates the strongest correlation across most aspects. We highlight the \textbf{highest} score in bold and the \underline{second-highest} score with underlines. }
\begin{adjustbox}{max width=\linewidth}
\begin{tabular}{lccccc}
\toprule
\textbf{Metric} & \textbf{COH} & \textbf{CON} & \textbf{FLU} & \textbf{REL} & \textbf{AVE} \\
\midrule 
\multicolumn{6}{l}{~~\textit{Rule-based metrics}} \\
BLEU & 0.110 & 0.071 & 0.050 & 0.248 & 0.120 \\
ROUGE & 0.172 & 0.142 & 0.113 & 0.270 & 0.174 \\
METEOR & 0.133 & 0.152 & 0.092 & 0.290 & 0.167 \\
\midrule
\multicolumn{6}{l}{~~\textit{Machine-learned metrics}} \\
BERTScore & 0.293 & 0.125 & 0.139 & 0.361 & 0.230\\
BARTScore & 0.140 & 0.101 & 0.112 & 0.293 & 0.162 \\
\midrule 
\multicolumn{6}{l}{~~\textit{Ours}} \\
LLMScore & 0.440 & \bf 0.366 & 0.362 & 0.277 & 0.361 \\ 
\hdashline
CoAScore$_{(n=5)}$ & \underline{0.499} & \underline{0.354} & \bf 0.379 & \underline{0.420} & \underline{0.413}\\ 
CoAScore$_{(n=10)}$ & \bf 0.541 & 0.339 & \underline{0.367} & \bf 0.478 & \bf 0.431\\ 
CoAScore$_{(n=20)}$ & 0.460 & 0.340 & 0.308 & 0.439 & 0.387\\ 
\bottomrule
\end{tabular}
\end{adjustbox}
\label{summarization_result}
\end{table}

\begin{table}[t]
\centering
\caption{Spearman correlations between metrics and human judgments on four aspects of the TopicalChat. CoAScore correlates best with human judgments across all these aspects. We highlight the \textbf{highest} score in bold and the \underline{second-highest} score with underlines. }
\begin{adjustbox}{max width=\linewidth}
\begin{tabular}{lccccc}
\toprule
\textbf{Metric} & \textbf{NAT} & \textbf{UND} & \textbf{INT} & \textbf{CON} & \textbf{AVE} \\
\midrule 
\multicolumn{6}{l}{~~\textit{Rule-based metrics}} \\
BLEU & 0.171 & 0.207 & 0.305 & 0.243 & 0.232 \\
ROUGE & 0.145 & 0.174 & 0.302 & 0.211 & 0.208\\
METEOR & 0.194 & 0.223 & 0.406 & 0.274 & 0.274 \\
\midrule
\multicolumn{6}{l}{~~\textit{Machine-learned metrics}} \\
BERTScore & 0.213 & 0.232 & 0.325 & 0.222 & 0.248\\
BARTScore & 0.213 & 0.207 & 0.436 & 0.275 & 0.283\\
\midrule 
\multicolumn{6}{l}{~~\textit{Ours}} \\
LLMScore & 0.514 & 0.410 & \underline{0.549} & 0.501 & 0.494 \\ 
\hdashline
CoAScore$_{(n=5)}$ & \underline{0.587} & \underline{0.531} & 0.546 & 0.544 & \underline{0.552} \\ 
CoAScore$_{(n=10)}$ & 0.571 & 0.530 & 0.507 & \underline{0.545} & 0.538 \\ 
CoAScore$_{(n=20)}$ & \bf 0.596 & \bf 0.542 & \bf 0.595 & \bf 0.553 & \bf 0.572 \\
\bottomrule
\end{tabular}
\end{adjustbox}
\label{dialog_result}
\end{table}

To demonstrate the generalizability of our approach, we extend our experiments beyond assessing the overall quality. 
Our investigations encompass two evaluation datasets, SummEval \cite{fabbri2021summeval} and TopicalChat \cite{mehri2020usr}, both with multi-aspect human-assessed scores.
Table \ref{summarization_result} presents the Spearman correlation between evaluation metrics and human judgments on the SummEval dataset. CoAScore demonstrates marked enhancement in \textbf{COH}erence, \textbf{FLU}ency, and \textbf{REL}evance aspects, with a slight decrease in \textbf{CON}sistency.
Simultaneously, Table \ref{dialog_result} showcases Spearman's correlation between various metrics and human judgments on the TopicalChat dataset.
CoAScore consistently improves the correlation across all aspects: \textbf{NAT}ural, \textbf{UND}erstandable, \textbf{INT}erest, and maintains \textbf{CON}text.
Across these two datasets, we calculate average correlation coefficients for multiple aspects respectively. We observe that CoAScore outperforms other unsupervised baselines by a large margin, regardless of whether 5, 10, or 20 aspects are considered.
Please refer to Appendix B for the comprehensive results of the multi-aspect assessments on the SummEval and TopicalChat datasets.

\begin{table}[t]
\centering
\caption{Effectiveness of the Relevant Aspect Generation stage. Utilizing the relevant aspects generated by LLM as references is preferable to directly employing other aspects already present in the evaluation dataset. We highlight the \textbf{highest} score in bold and the \underline{second-highest} score with underlines. ``\textcolor{red}{Red}'' denotes CoAScore$_{inter}$ is better than LLMScore while ``\textcolor{blue}{Blue}'' denotes the opposite.}
\begin{adjustbox}{max width=\linewidth}
\begin{tabular}{lccccc}
\toprule
\textbf{Metric} & \textbf{NAT} & \textbf{UND} & \textbf{INT} & \textbf{CON} & \textbf{OVE} \\
\midrule 
LLMScore & 0.514 & 0.410 & 0.549 & 0.501 & 0.592 \\ 
CoAScore$_{inter}$ & \textcolor{blue}{0.494} & \textcolor{red}{0.500} & \underline{\textcolor{red}{0.566}} & \textcolor{blue}{0.399} & \textcolor{red}{0.628} \\
CoAScore$_{(n=5)}$ & \underline{0.587} & \underline{0.531} & 0.546 & 0.544 & 0.625 \\ 
CoAScore$_{(n=10)}$ & 0.571 & 0.530 & 0.507 & \underline{0.545} & \underline{0.638} \\ 
CoAScore$_{(n=20)}$ & \bf 0.596 & \bf 0.542 & \bf 0.595 & \bf 0.553 & \bf 0.641 \\
\bottomrule
\end{tabular}
\end{adjustbox}
\label{stage1_effect}
\end{table}

\begin{figure}[t]
\centering
\includegraphics[width=0.45\textwidth]{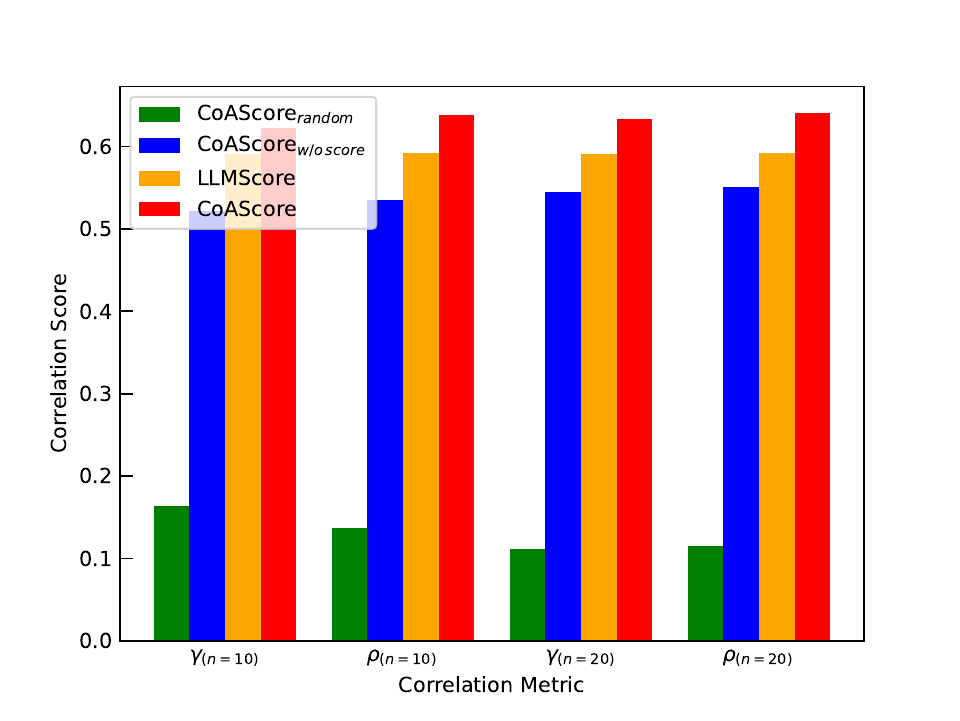}
\caption{Effectiveness of the Relevant Aspect Scoring stage. Owing to the absence of reference scores, the performance of CoAScore$_{w/o\,score}$ falls behind that of LLMScore. Furthermore, assigning random scores to relevant aspects seriously distort the evaluation of a specific aspect, resulting in the weakest correlation scores of CoAScore$_{random}$.}
\label{ablation_score}
\end{figure}

\begin{figure*}[htbp]
\centering
\includegraphics[width=\textwidth]{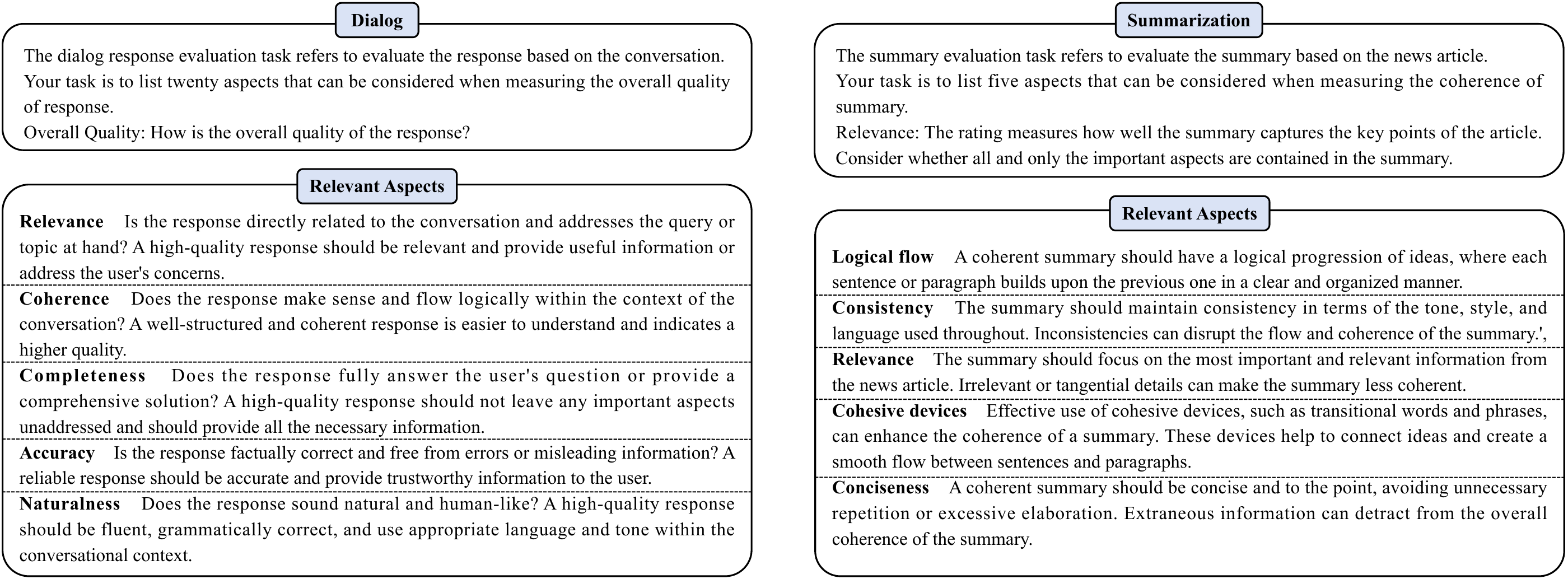}
\caption{
    Examples of Relevant Aspect Generation in evaluating the overall quality of dialogue responses and the coherence of summaries. Each one provides five relevant aspects as references to help the target aspect evaluation.
}
\label{case}
\end{figure*}

\begin{table}[t]
\centering
\caption{Effectiveness of the Chain-of-Aspects Scoring stage. 
LLM improves evaluation accuracy by utilizing relevant aspect scores instead of directly averaging various scores. CoAScore$_{average}$ vs CoAScore: \colorbox{red!10}{Red} denotes better, \colorbox{blue!10}{Blue} signifies worse. \textbf{AC} means the count of relevant aspects.}
\begin{adjustbox}{max width=\linewidth}
\begin{tabular}{lcccc}
\toprule
\multirow{2}{*}{\textbf{Metric}} & \multirow{2}{*}{\textbf{AC}} & \multicolumn{3}{c}{\textbf{OpenMEVA}} \\
\cmidrule(lr){3-5} 
& & $\gamma$ & $\rho$ & $\tau$ \\
\midrule 
CoAScore$_{average}$ & 5 & \cellcolor{red!10}0.316 & \cellcolor{blue!10}0.315 & \cellcolor{red!10}0.238\\ 
CoAScore & 5 & \cellcolor{blue!10}0.295 & \cellcolor{red!10}0.316 & \cellcolor{blue!10}0.237\\ 
\hdashline
CoAScore$_{average}$ & 10 & \cellcolor{blue!10}0.388 & \cellcolor{blue!10}0.410 & \cellcolor{blue!10}0.300\\ 
CoAScore & 10 & \cellcolor{red!10}0.408 & \cellcolor{red!10}0.414 & \cellcolor{red!10}0.304\\ 
\hdashline
CoAScore$_{average}$ & 20 & \cellcolor{blue!10}0.402 & \cellcolor{blue!10}0.401 & \cellcolor{blue!10}0.288\\ 
CoAScore & 20 & \cellcolor{red!10}0.414 & \cellcolor{red!10}0.411 & \cellcolor{red!10}0.315\\ 
\bottomrule
\end{tabular}
\end{adjustbox}
\label{openmeva_stage3}
\end{table}

\begin{figure}[t]
\centering
\includegraphics[width=0.45\textwidth]{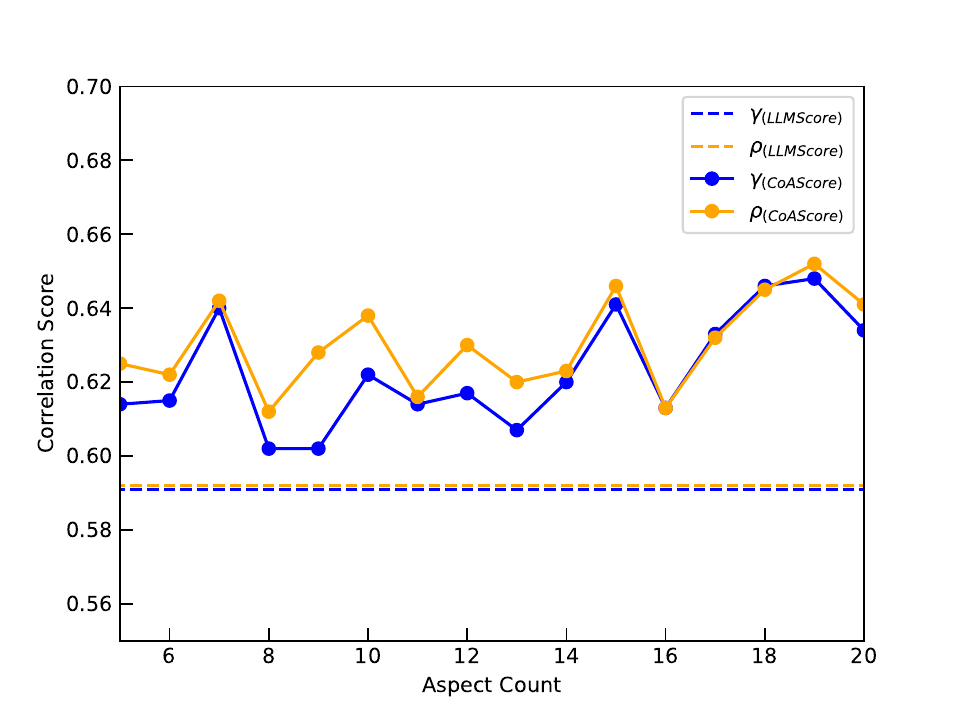}
\caption{
    Effectiveness of vairous relevant aspect numbers. As the number of relevant aspects increases, the correlation scores of CoAScore is generally improved and always better than the ones of LLMScore.
}
\label{aspect_count}
\end{figure}

\begin{table*}[t]
    \centering
        \caption{Example from the TopicalChat dataset. \textcolor{red}{'Red'} denotes the better correlation whereas \textcolor{blue}{'Blue'} denotes the wrong one. CoAScore scores align with human judgments.}
    \begin{tabularx}{\linewidth}{X}
        \toprule
        \textbf{Conversation} \\
        \midrule
        A: Hi , how are you? \\
        B: I am good, thank you. I enjoy travel. Do you?\\
        A: I do but I am a worrying traveler. Where is the best place you have been to?\\
        B: I have not been out of the country but several places in the country. I like chicago a lot. Did you know that all of japans highways are tolls?\\
        % A: That's got ta be expensive! what's the best place in chicago? where should I go when i visit?\\
        % B: I like the natural history museum and millennium park. Lou malnati's has amazing pizza. It is very expensive to travel in japan. It costs more than \$300 to travel across the country.\\
        ...\\
        A: Can you imagine the movie speed on that bus!!!! holy smokes!\\
        B: Ha ha, yes that would change that movie quite a bit. Back to iceland , most don't have cars and travel between towns by flight!\\
        A: Really. Seems a little overkill but what do i know... I want to go there and see the northern lights some day.\\
        B: That would be a beautiful sight. Off topic but how do you feel about sharks?\\
        A: I'm for em! \\
        B: They are very interesting. Did you know that they don't have rib cages?\\
        \midrule
        \textbf{Response} \\
        \midrule
        Wow really? So their heart is unprotected?\\
        \midrule
        \textbf{Relevant Aspect Scores} \\
        \midrule
        \textbf{Relevance}: 5.0 \hspace{1.6cm} \textbf{Coherence}: 5.0 \hspace{1.6cm} \textbf{Completeness}: 4.0 \hspace{1.6cm} \textbf{Accuracy}: 5.0 \hspace{1.6cm} \textbf{Naturalness}: 5.0 \\
        \midrule
        \textbf{Overll Quality Score} \\
        \midrule
        \textbf{LLMScore}: \textcolor{blue}{3.0} \hspace{5.9cm} \textbf{CoAScore}: \textcolor{red}{5.0} \hspace{5.9cm} \textbf{Human}: 5.0 \\
        \bottomrule
    \end{tabularx}
    \label{dialog_example}
\end{table*}

\subsection{Analyses}
\paragraph{Effect of Relevant Aspect Generation}
To illustrate why we employ LLM to generate relevant aspects and demonstrate its effectiveness within the CoAScore framework, we compared two methods: direct aspect generation through LLM and selecting from the evaluation dataset (CoAScore$_{inter}$). We experimented with five aspects in the TopicalChat dataset \cite{mehri2020usr}, reporting Spearman correlations in Table \ref{stage1_effect}.
Our findings show that CoAScore$_{inter}$ exhibits lower correlation scores for several aspects (e.g, Natural and Maintains Context) compared to CoAScore, possibly due to a lack of consideration for interrelationships between different aspects during dataset construction \cite{mehri2020usr}. Furthermore, CoAScore$_{inter}$ shows weaker correlations with human assessments across all aspect evaluations compared to CoAScore. 
This highlights that LLM-generated relevant aspects are more conducive to LLM itself in making evaluation judgments. 
Additionally, as CoAScore can automatically generate various relevant aspects, the differences between the two approaches become more pronounced with an increasing number of relevant aspects, underscoring the significance of integrating LLM-generated relevant aspects into the CoAScore framework.

\paragraph{Effect of Relevant Aspect Scoring} 
To test whether the scores of relevant aspects do help the evaluation of the target aspect, we conduct an additional experiment with two variants of CoAScore, compare both with LLMScore and CoAScore in evaluating the overall quality of the TopicalChat dataset \cite{mehri2020usr}, measured by Pearson \cite{mukaka2012guide} and Spearman \cite{zar2005spearman}, as shown in Figure \ref{ablation_score}.
The CoAScore without reference scores is termed CoAScore$_{w/o\,score}$. We find that without scores, CoAScore$_{w/o \, score}$ performs lower than LLMScore, proving the importance of reference scores for CoAScore.
The CoAScore with randomized reference scores is termed CoAScore$_{random}$. We observe that CoAScore$_{random}$ correlates worst with human judgments, which means that randomly assigned scores of relevant aspects can even mislead LLM-based evaluation assessments, resulting in diminished correlation with human judgments. 
Furthermore, as the number of relevant aspects increases (from 10 to 20), introducing random perturbations to aspect scores exacerbates the decline in correlation with manual evaluation. This underscores the scores of relevant aspects do play an important role in the evaluation and it is crucial to derive accurate reference scores in the proposed CoAScore framework.

\paragraph{Effect of Chain-of-Aspects Scoring}
To assess the need for LLM to reference relevant aspect scores for final evaluation results, we devised an alternative approach, CoAScore$_{average}$, which directly averages multiple relevant aspect scores, omitting the "chain-of-aspect scoring" in CoAScore (Stage 3).
We compared CoAScore and CoAScore$_{average}$ across 5, 10, and 20 aspects in the OpenMEVA dataset \cite{guan2021openmeva}, reporting Pearson, Spearman, and Kendall-Tau correlation scores in Table \ref{openmeva_stage3}.
In most cases, CoAScore achieves higher correlation scores than CoAScore$_{average}$, reinforcing the effectiveness of LLM incorporating various relevant aspect scores for assessments. Notably, as the number of relevant aspects increased, CoAScore's advantage over CoAScore${average}$ in correlation scores became more pronounced, highlighting the importance of LLM incorporating these scores into its deliberations for evaluation, beyond simple averaging.

\paragraph{Effect of Relevant Aspect Count}
To demonstrate the robustness of CoAScore concerning the count of relevant aspects, we vary the number of aspects (ranging from 5 to 20) in computing the CoAScore. We experiment on the overall quality of the TopicalChat dataset \cite{mehri2020usr}, using Pearson and Spearman correlation measures \cite{mukaka2012guide, zar2005spearman}.
Experimental results in Figure \ref{aspect_count} show that irrespective of the number of aspects, the correlation scores between CoAScore and human judgments consistently surpass those of the LLMScore. This finding demonstrates the robustness of CoAScore regarding the number of relevant aspects.
Moreover, our findings indicate that as the number of relevant aspects increases, although minor performance fluctuations occur within different intervals, the overall correlation slightly increases, which indicates a potential benefit of leveraging more comprehensive knowledge. 

\paragraph{Case Study}
In Figure \ref{case}, we present two cases of the generated chain of aspects. 
For evaluating the overall quality of dialogue responses \cite{mehri2020usr}, CoAScore generates five relevant aspects of overall quality, namely relevance, coherence, completeness, accuracy, and naturalness, along with their corresponding descriptions. These aspects demonstrate a high correlation with overall quality. CoAScore also aids in assessing coherence in summaries \cite{fabbri2021summeval} by generating relevant aspects like logical flow, consistency, relevance, cohesive devices, and conciseness, all strongly aligned with coherence. The LLM-generated aspects are found to be sensible and task-appropriate. These aspects largely contribute to the evaluation process, highlighting the reliable effectiveness of LLM-generated relevant aspects.

In our analysis, when there is a substantial deviation between LLMScore and human judgments in evaluating a specific aspect, CoAScore's evaluation of relevant aspects serves as a robust reference framework. This aids in the assessment of the target aspect, reducing disparities with human judgments.
As illustrated in Table \ref{dialog_example}, it becomes evident that while humans award a high rating to a top-quality response, LLMScore assigns a considerably lower score. This incongruity highlights the limitations of LLMScore in appraising excellent responses.
In contrast, CoAScore closely aligns with human judgments. It accomplishes this by initially generating a chain of relevant aspects, assigning scores to each aspect, and utilizing relevant aspect scores to gauge the target aspect. CoAScore's attribution of high scores to all relevant aspects contributes to a higher rating for the "Overall" aspect, thus modifying the initial outcome.
For additional specific examples and detailed information, please consult Appendix C.

\section{Conclusion and future work}
In this research, we present a novel LLM-based evaluation metric called CoAScore. Unlike conventional metrics that assess hypotheses' aspects individually, CoAScore generates a chain of relevant aspects for the target aspect, initially assigns scores to them, and subsequently employs this knowledge, including descriptions and scores, as references to improve the evaluation of the given aspect.
Our experiments reveal that CoAScore exhibits higher correlations with human judgments than a range of unsupervised evaluation metrics. This holds true for overall quality as well as specific aspects when compared to isolated evaluations of individual aspects.

For future work, we will explore innovative strategies to improve the efficiency of CoAScore evaluations. This includes utilizing larger models to guide smaller models in scoring relevant aspects and harnessing the knowledge of relevant aspects for target aspect assessments. Additionally, we will investigate CoAScore's applicability in specific NLG evaluation tasks, such as assessing hallucinations in LLM outputs.

\bibliographystyle{ACM-Reference-Format}
\bibliography{sample-base}

\newpage

\appendix

\section{Appendix}

\subsection{Detailed CoAScore Prompts}
CoAScore leverages a chain-of-aspects prompting framework to employ the knowledge of relevant aspects to measure the given aspect of hypotheses, consisting of three core components. We illustrate the LLM-based baseline metric and each component of CoAScore with the dialog response evaluation task:

\paragraph{LLMScore} Our self-designed LLM-based evaluation metric, which is used to compare with CoAScore, as shown in Table \ref{llmscore}.

\paragraph{Relevant Aspect Generation} Before evaluating a given aspect, CoAScore preferentially generates a chain of aspects related to the target aspect as references for the final evaluation, as depicted in Table \ref{rag}.

\paragraph{Relevant Aspect Scoring} After obtaining the relevant aspects, CoAScore scores the quality of the response in each aspect, as illustrated in Table \ref{ras}.

\paragraph{Chain-of-Aspects Scoring} CoAScore leverages the knowledge (descriptions and scores) of relevant aspects to enhance the ability to evaluate the target aspect, as illustrated in Table \ref{coa}.

\subsection{Detailed Results on Multi-Aspect Datasets}
Compared to Table \ref{summarization_result} and Table \ref{dialog_result}, we list comprehensive results (Pearson, Spearman, Kendall-Tau) of our metrics and unsupervised baseline metrics on multi-aspect evaluation datasets for the following:

\paragraph{SummEval} Table \ref{summeval_results2} demonstrates that CoAScore exhibits stronger alignment with human assessments across the majority of aspects, with only a marginal decrease observed in the case of the consistency aspect, whether with 5, 10, or 20 relevant aspects.

\paragraph{TopicalChat} Table \ref{Topicalchat_results2} showcases that CoAScore consistently shows a better alignment with human judgments across all aspects, regardless of the number of relevant aspects (5, 10, or 20).

\subsection{Cases}
% We propose two examples of different NLG evaluation tasks to verify why CoAScore correlates better with human judgments than LLM-based baseline metric. 
% Table \ref{dialog} illustrates that for evaluating the overall quality of the high-quality response, human annotators propose a high score while the score of LLMScore is much lower. Therefore, LLMScore is not accurate for high-quality response evaluation. Meanwhile, CoAScore scores align with human judgments due to before scoring the overall quality directly, it thinks in terms of multiple relevant aspects and evaluates the score for each one, leveraging the knowledge of all relevant aspects to evaluate the target aspect.
% Table \ref{summ} demonstrates then measuring the coherence of the high-quality summary, the score of LLMScore is much lower than the one of humans, leading a wrong judgment for this case. For CoAScore, it measures the qualities of relevant aspects for the summary first. We fine CoAScore proposes the summary a high score for relevance, logical flow and conciseness aspect, and lead to a higher score than LLMScore, which correlates better with human judgments.
We provide two distinct examples of different NLG evaluation tasks to validate the superior correlation between CoAScore and human judgments compared to the LLM-based baseline metric.

In Table \ref{dialog}, it's evident that when assessing the overall quality of a high-quality response, human evaluators assign a high score, whereas LLMScore's score is considerably lower. This discrepancy highlights LLMScore's inaccuracy in evaluating high-quality responses. Conversely, CoAScore aligns more closely with human judgments. It achieves this by initially scoring multiple relevant aspects and leveraging its knowledge to evaluate the target aspect, thus enhancing its overall quality evaluation.

Moving to Table \ref{summ}, when measuring the coherence of a high-quality summary, LLMScore's score is significantly lower than the human-assigned score, leading to an erroneous judgment. CoAScore, however, scores the relevant aspects of the summary first, where it particularly excels in relevance, logical flow, and conciseness. This results in a higher score than LLMScore and demonstrates better alignment with human judgments.

\begin{table*}[htbp]
    \centering
        \caption{LLMScore Prompt on the dialog response evaluation task. The clear task description are in \textbf{bold}}
    \begin{tabularx}{\linewidth}{X}
        \toprule
        \textbf{Input} \\
        \midrule
        You will be given a conversation between two individuals.\\
        You will then be given one potential response for the next turn in the conversation.\\
        The response concerns an interesting fact, which will be provided as well.\\
        \textbf{Your task is to rate the response on one metric.} \\
        \\
        Evaluation Criteria: \\
        Overall Quality (1-5) How is the overall quality of the response?\\
        \\
        Fact:\\
        Bill murray thinks kung fu hustle is "the supreme achievement of the modern age in terms of comedy".\\
        \\
        Conversation: \\
        A: Morning ! are you a comedy fan? \\
        B: Morning ! yes, some of it. Have you watched anything good recently? \\
        A: Kung fu hustle . ever seen it? It 's hilarious.\\
        B: i haven't ! what is it about? Does it have bruce lee in it?\\
        A: No, it's much more recent. Steven chow directed it. Bill murray loved it too.\\
        \\
        Response:\\
        Yeah bill murray called it "the supreme achievement of the modern age in terms of comedy"\\
        \\
        \textbf{Based on the conversation and the evaluation criteria for overall quality, please rate the overall quality of the response.}\\
        Overall Quality Score:\\
        \midrule
        \textbf{Output} \\
        \midrule
        4.0\\
        \bottomrule
    \end{tabularx}
    \label{llmscore}
\end{table*}

\begin{table*}[htbp]
    \centering
        \caption{Relevant Aspect Generation Prompt on the dialog response evaluation task. Five relevant aspects are generated for the overall quality aspect. The clear task description and aspect names are in \textbf{bold}.}
    \begin{tabularx}{\linewidth}{X}
        \toprule
        \textbf{Input} \\
        \midrule
        The dialog response evaluation task refers to evaluate the response based on the conversation. \\
        \textbf{Your task is to list five aspects that can be considered when measuring the overall quality of response.} \\
        Overall Quality: How is the overall quality of the response?\\
        \midrule
        \textbf{Output} \\
        \midrule
        \textbf{Relevance}: Is the response directly related to the conversation and addresses the query or topic at hand? A high-quality response should be relevant and provide useful information or address the user's concerns.\\
        \textbf{Coherence}: Does the response make sense and flow logically within the context of the conversation? A well-structured and coherent response is easier to understand and indicates a higher quality. \\
        \textbf{Completeness}: Does the response fully answer the user's question or provide a comprehensive solution? A high-quality response should not leave any important aspects unaddressed and should provide all the necessary information.\\
        \textbf{Accuracy}: Is the response factually correct and free from errors or misleading information? A reliable response should be accurate and provide trustworthy information to the user.\\
        \textbf{Naturalness}: Does the response sound natural and human-like? A high-quality response should be fluent, grammatically correct, and use appropriate language and tone within the conversational context.\\
        \bottomrule
    \end{tabularx}
    \label{rag}
\end{table*}

\begin{table*}[htbp]
    \centering
    \caption{Relevant Aspect Scoring Prompt on the dialog response evaluation task. Scoring each relevant aspect for the response. The clear task description and aspect names are in \textbf{bold}.}
    \begin{tabularx}{\linewidth}{X}
        \toprule
        \textbf{Input} \\
        \midrule
        You will be given a conversation between two individuals.\\
        You will then be given one potential response for the next turn in the conversation.\\
        The response concerns an interesting fact, which will be provided as well.\\
        \textbf{Your task is to rate the response based on the given aspects.} \\
        Scores for each aspect range from 1 to 5, representing worst to best. \\
        \\
        Aspects: \\
        \textbf{Relevance}: Is the response directly related to the conversation and addresses the query or topic at hand? A high-quality response should be relevant and provide useful information or address the user's concerns.\\
        \textbf{Coherence}: Does the response make sense and flow logically within the context of the conversation? A well-structured and coherent response is easier to understand and indicates a higher quality. \\
        \textbf{Completeness}: Does the response fully answer the user's question or provide a comprehensive solution? A high-quality response should not leave any important aspects unaddressed and should provide all the necessary information.\\
        \textbf{Accuracy}: Is the response factually correct and free from errors or misleading information? A reliable response should be accurate and provide trustworthy information to the user.\\
        \textbf{Naturalness}: Does the response sound natural and human-like? A high-quality response should be fluent, grammatically correct, and use appropriate language and tone within the conversational context.\\
        \\
        Fact:\\
        Bill murray thinks kung fu hustle is "the supreme achievement of the modern age in terms of comedy".\\
        \\
        Conversation: \\
        A: Morning ! are you a comedy fan? \\
        B: Morning ! yes, some of it. Have you watched anything good recently? \\
        A: Kung fu hustle . ever seen it? It 's hilarious.\\
        B: i haven't ! what is it about? Does it have bruce lee in it?\\
        A: No, it's much more recent. Steven chow directed it. Bill murray loved it too.\\
        \\
        Response:\\
        Yeah bill murray called it "the supreme achievement of the modern age in terms of comedy"\\
        \\
        \textbf{Based on the conversation and the aspects, please rate the response for each aspect.}\\
        Provide them in JSON format, aspect as key, score as value:\\
        \midrule
        \textbf{Output} \\
        \midrule
        \textbf{Relevance}: 5.0\\
        \textbf{Coherence}: 5.0\\
        \textbf{Completeness}: 4.0\\
        \textbf{Accuracy}: 5.0\\
        \textbf{Naturalness}: 5.0\\
        \bottomrule
    \end{tabularx}

    \label{ras}
\end{table*}

\begin{table*}[htbp]
    \centering
        \caption{Chain-of-Aspects Scoring Prompt on the dialog response evaluation task. Leveraging the knowledge of relevant aspects to evaluate the target aspect of the response. The clear task description and aspect names are in \textbf{bold}.}
    \begin{tabularx}{\linewidth}{X}
        \toprule
        \textbf{Input} \\
        \midrule
        You will be given a conversation between two individuals.\\
        You will then be given one potential response for the next turn in the conversation.\\
        The response concerns an interesting fact, which will be provided as well.\\
        \textbf{Your task is to rate the response on one metric.} \\
        \\
        Evaluation Criteria: \\
        Overall Quality (1-5) How is the overall quality of the response?\\
        \\
        Fact:\\
        Bill murray thinks kung fu hustle is "the supreme achievement of the modern age in terms of comedy".\\
        \\
        Conversation: \\
        A: Morning ! are you a comedy fan? \\
        B: Morning ! yes, some of it. Have you watched anything good recently? \\
        A: Kung fu hustle . ever seen it? It 's hilarious.\\
        B: i haven't ! what is it about? Does it have bruce lee in it?\\
        A: No, it's much more recent. Steven chow directed it. Bill murray loved it too.\\
        \\
        Response:\\
        Yeah bill murray called it "the supreme achievement of the modern age in terms of comedy"\\
        \\
        Before you rate the above response, some scores for different aspects of this response can help you rate the overall quality of this response:\\
        \textbf{Relevance}: Is the response directly related to the conversation and addresses the query or topic at hand? A high-quality response should be relevant and provide useful information or address the user's concerns.\\
        Score: 5.0\\
        \textbf{Coherence}: Does the response make sense and flow logically within the context of the conversation? A well-structured and coherent response is easier to understand and indicates a higher quality. \\
        Score: 5.0\\
        \textbf{Completeness}: Does the response fully answer the user's question or provide a comprehensive solution? A high-quality response should not leave any important aspects unaddressed and should provide all the necessary information.\\
        Score: 4.0\\
        \textbf{Accuracy}: Is the response factually correct and free from errors or misleading information? A reliable response should be accurate and provide trustworthy information to the user.\\
        Score: 5.0\\
        \textbf{Naturalness}: Does the response sound natural and human-like? A high-quality response should be fluent, grammatically correct, and use appropriate language and tone within the conversational context.\\
        Score: 5.0\\
        \\
        \textbf{Based on the conversation, the evaluation criteria for overall quality and the scores for different aspects of the response, please rate the overall quality of the response.}\\
        Overall Quality Score:\\
        \midrule
        \textbf{Output} \\
        \midrule
        5.0\\
        \bottomrule
    \end{tabularx}
    \label{coa}
\end{table*}

\begin{table*}[htbp]
\centering
\caption{Correlations between metrics and human judgments regarding various aspects on the SummEval dataset. Our proposed CoAScore exhibits stronger correlations with human judgments across most aspects. We highlight the \textbf{highest} score in bold and the \underline{second-highest} score with underlines. }
\begin{adjustbox}{max width=\linewidth}
\begin{tabular}{lcccccccccccc}
\toprule
\multirow{2}{*}{\textbf{Metric}} &
\multicolumn{3}{c}{\textbf{Coherence}} & \multicolumn{3}{c}{\textbf{Consistency}} &
\multicolumn{3}{c}{\textbf{Fluency}} & \multicolumn{3}{c}{\textbf{Relevance}} \\
\cmidrule(lr){2-4} \cmidrule(lr){5-7} \cmidrule(lr){8-10} \cmidrule(lr){11-13}  
&  $\gamma$ & $\rho$ & $\tau$ & $\gamma$ & $\rho$ & $\tau$ & $\gamma$ & $\rho$ & $\tau$ & $\gamma$ & $\rho$ & $\tau$ \\
\midrule 
\multicolumn{10}{l}{~~\textit{Rule-based metrics}} \\
BLEU & 0.115 & 0.110 & 0.077 & 0.076 & 0.071 & 0.056 & 0.072 & 0.050 & 0.039 & 0.205 & 0.248 & 0.177\\
ROUGE & 0.183 & 0.172 & 0.121 & 0.159 & 0.142 & 0.112 & 0.122 & 0.113 & 0.087 & 0.268 & 0.270 & 0.193\\
METEOR & 0.142 & 0.133 & 0.095 & 0.180 & 0.152 & 0.119 & 0.117 & 0.092 & 0.071 & 0.305 & 0.290 & 0.209\\
\midrule
\multicolumn{10}{l}{~~\textit{Machine-learned metrics}} \\
BERTScore & 0.308 & 0.293 & 0.207 & 0.151 & 0.125 & 0.098 & 0.182 & 0.139 & 0.108 & 0.370 & 0.362 & 0.261\\
BARTScore& 0.120 & 0.140 & 0.100 & 0.110 & 0.101 & 0.079 & 0.118 & 0.112 & 0.088 & 0.284 & 0.293 & 0.209\\
\midrule 
\multicolumn{10}{l}{~~\textit{Ours}} \\
LLMScore & 0.449 & 0.440 & 0.371 & \bf 0.479 & \bf 0.366 & \bf 0.345 & 0.480 & 0.362 & \underline{0.338} & 0.334 & 0.277 & 0.236\\
\hdashline
CoAScore$_{(n=5)}$ & \underline{0.491} & \underline{0.499} & \underline{0.397} & 0.447 & \underline{0.354} & \underline{0.303} & \underline{0.481} & \bf 0.379 & \bf 0.340 & 0.450 & 0.420 & 0.339 \\
CoAScore$_{(n=10)}$ & \bf 0.526 & \bf 0.541 & \bf 0.419 & 0.416 & 0.339 & 0.299 & \bf 0.485 & \underline{0.367} & 0.308 & \bf 0.494 & \bf 0.478 & \bf 0.379\\
CoAScore$_{(n=20)}$ & 0.466 & 0.460 & 0.366 & \underline{0.462} & 0.340 & 0.297 & 0.471 & 0.308 & 0.262 & \underline{0.451} & \underline{0.439} & \underline{0.368} \\
\bottomrule
\end{tabular}
\end{adjustbox}
\label{summeval_results2}
\end{table*}

\begin{table*}[htbp]
\centering
\caption{Correlations between metrics and human judgments regarding various aspects on the TopicalChat dataset. Our proposed CoAScore exhibits stronger correlations with human judgments across all these aspects. We highlight the \textbf{highest} score in bold and the \underline{second-highest} score with underlines. }
\begin{adjustbox}{max width=\linewidth}
\begin{tabular}{lcccccccccccc}
\toprule
\multirow{2}{*}{\textbf{Metric}} &
\multicolumn{3}{c}{\textbf{Natural}} & \multicolumn{3}{c}{\textbf{Understandable}} &
\multicolumn{3}{c}{\textbf{Interest}} & \multicolumn{3}{c}{\textbf{Context}} \\
\cmidrule(lr){2-4} \cmidrule(lr){5-7} \cmidrule(lr){8-10} \cmidrule(lr){11-13}  
&  $\gamma$ & $\rho$ & $\tau$ & $\gamma$ & $\rho$ & $\tau$ & $\gamma$ & $\rho$ & $\tau$ & $\gamma$ & $\rho$ & $\tau$ \\
\midrule 
\multicolumn{10}{l}{~~\textit{Rule-based metrics}} \\
BLEU & 0.184 & 0.171 & 0.123 & 0.212 & 0.207 & 0.156 & 0.242 & 0.305 & 0.217 & 0.153 & 0.243 & 0.175\\
ROUGE & 0.170 & 0.145 & 0.103 & 0.214 & 0.174 & 0.133 & 0.293 & 0.302 & 0.221 & 0.193 & 0.211 & 0.151 \\
METEOR & 0.222 & 0.194 & 0.137 & 0.258 & 0.223 & 0.171 & 0.399 & 0.406 & 0.293 & 0.266 & 0.274 & 0.195 \\
\midrule
\multicolumn{10}{l}{~~\textit{Machine-learned metrics}} \\
BERTScore & 0.226 & 0.213 & 0.152 & 0.260 & 0.232 & 0.176 & 0.311 & 0.325 & 0.233 & 0.205 & 0.222 & 0.154\\
BARTScore & 0.213 & 0.213 & 0.148 & 0.220 & 0.207 & 0.153 & 0.412 & 0.436 & 0.308 & 0.261 & 0.275 & 0.192\\
\midrule 
\multicolumn{10}{l}{~~\textit{Ours}} \\
LLMScore & 0.493 & 0.514 & 0.426 & 0.387 & 0.410 & 0.367 & \underline{0.548} & \underline{0.549} & \underline{0.477} & 0.491 & 0.501 & 0.426 \\
\hdashline
CoAScore$_{(n=5)}$ & \bf 0.567 & \underline{0.587} & \underline{0.477} & \underline{0.409} & \underline{0.531} & \underline{0.434} & 0.502 & 0.546 & 0.439 & \underline{0.534} & 0.544 & 0.426\\
CoAScore$_{(n=10)}$ & 0.541 & 0.571 & 0.457 & 0.405 & 0.530 & 0.432 & 0.432 & 0.507 & 0.420 & 0.530 & \underline{0.545} & \underline{0.434} \\
CoAScore$_{(n=20)}$ & \underline{0.558} & \bf 0.596 & \bf 0.482 & \bf 0.441 & \bf 0.542 & \bf 0.458 & \bf 0.578 & \bf 0.595 & \bf 0.491 & \bf 0.539 & \bf 0.553 & \bf 0.441 \\
\bottomrule
\end{tabular}
\end{adjustbox}
\label{Topicalchat_results2}
\end{table*}

\begin{table*}[htbp]
    \centering
        \caption{Example from the TopicalChat dataset. \textcolor{red}{'Red'} denotes the better correlation whereas \textcolor{blue}{'Blue'} denotes the wrong one. CoAScore scores align with human judgments.}
    \begin{tabularx}{\linewidth}{X}
        \toprule
        \textbf{Conversation} \\
        \midrule
        A: Hi , how are you? \\
        B: I am good, thank you. I enjoy travel. Do you?\\
        A: I do but I am a worrying traveler. Where is the best place you have been to?\\
        B: I have not been out of the country but several places in the country. I like chicago a lot. Did you know that all of japans highways are tolls?\\
        A: That's got ta be expensive! what's the best place in chicago? where should I go when i visit?\\
        B: I like the natural history museum and millennium park. Lou malnati's has amazing pizza. It is very expensive to travel in japan. It costs more than \$300 to travel across the country.\\
        A: That's a lot. I can not imagine having to pay that in tolls. I do like lou 's pizza. Have you been to wrigley field? I should ask first if you are a baseball fan ...\\
        B: I have not. We did stop outside of soldier field but didn't get to go in. Did you know that iceland has no public rail service? \\
        A: That's odd. Soldier field is just ok. The shedd is really good though. where else, besides chicago do you like going?\\
        B: I enjoyed the mountains of tennessee. They are beautiful. We got a cabin with an amazing view. I would go to a lot more places if you could get there faster. The dutch are working on that. They are creating a commuter bus that goes 160 mph. That would be great.\\
        A: Can you imagine the movie speed on that bus!!!! holy smokes!\\
        B: Ha ha, yes that would change that movie quite a bit. Back to iceland , most don't have cars and travel between towns by flight!\\
        A: Really. Seems a little overkill but what do i know... I want to go there and see the northern lights some day.\\
        B: That would be a beautiful sight. Off topic but how do you feel about sharks?\\
        A: I'm for em! \\
        B: They are very interesting. Did you know that they don't have rib cages?\\
        \midrule
        \textbf{Response} \\
        \midrule
        Wow really? So their heart is unprotected?\\
        \midrule
        \textbf{Relevant Aspect Scores} \\
        \midrule
        \textbf{Relevance}: 5.0 \quad\quad\quad\quad \textbf{Coherence}: 5.0 \quad\quad\quad\quad \textbf{Completeness}: 4.0 \quad\quad\quad\quad \textbf{Accuracy}: 5.0 \quad\quad\quad\quad \textbf{Naturalness}: 5.0 \\
        \midrule
        \textbf{Overll Quality Score} \\
        \midrule
        \textbf{LLMScore}: \textcolor{blue}{3.0} \hspace{5.4cm} \textbf{CoAScore}: \textcolor{red}{5.0} \hspace{5.4cm} \textbf{Human}: 5.0 \\
        \bottomrule
    \end{tabularx}
    \label{dialog}
\end{table*}

\begin{table*}[htbp]
    \centering
        \caption{Example from the SummEval dataset. \textcolor{red}{'Red'} denotes the better correlation whereas \textcolor{blue}{'Blue'} denotes the wrong one. CoAScore correlates better than LLMScore.}
    \begin{tabularx}{\linewidth}{X}
        \toprule
        \textbf{News Article} \\
        \midrule
        Manchester City are keen to sign Anderlecht teenager Evangelos Patoulidis. The 14-year-old playmaker is regarded as one of the best talents to emerge from Anderlecht's youth set-up and has also attracted attention from Arsenal and Barcelona. The Belgian starlet rejected a move to Barcelona's La Masia academy when he was 12 as his family wanted him to continue his studies. He has continued to impress and City have held discussions with Anderlecht chairman Roger Vanden Stock in the hope of agreeing a compensation package. Manuel Pellegrini is looked to build for the future by snapping up hot property Evangelos Patoulidis.\\
        \midrule
        \textbf{Summary} \\
        \midrule
        Anderlecht are keen to sign Anderlecht teenager Evangelos Patoulidis. The 14-year-old playmaker is regarded as one of the best talents to emerge from Anderlecht's youth set-up and has also attracted attention from Arsenal and Barcelona. The Belgian starlet rejected a move to Barcelona's La Masia academy when he was 12 as his family wanted him to continue his studies.\\
        \midrule
        \textbf{Relevant Aspect Scores} \\
        \midrule
        \textbf{Logical flow}: 4.0 \hspace{1cm} \textbf{Consistency}: 3.0 \hspace{1cm} \textbf{Relevance}: 5.0 \hspace{1cm} \textbf{Cohesive devices}: 3.0 \hspace{1.05cm} \textbf{Conciseness}: 4.0 \\
        \midrule
        \textbf{Overll Quality Score} \\
        \midrule
        \textbf{LLMScore}: \textcolor{blue}{2.0} \hspace{5.4cm} \textbf{CoAScore}: \textcolor{red}{4.0} \hspace{5.4cm} \textbf{Human}: 5.0 \\
        \bottomrule
    \end{tabularx}
    \label{summ}
\end{table*}

\end{document}